%% file: main.tex
\documentclass[runningheads]{llncs}

\usepackage[final,year=2026]{eccv}

\usepackage{eccvabbrv}

\usepackage{graphicx}
\usepackage{booktabs}

\usepackage[accsupp]{axessibility}  % Improves PDF readability for those with disabilities.
\ifdefined\pdfcompresslevel
\fi
\usepackage{multirow}        % \multirow
\usepackage[table]{xcolor}   % \cellcolor

\providecommand{\nrgb}{Non-RGB}
\providecommand{\rgb}{RGB}
\providecommand{\avg}{Avg}
\usepackage[pdfversion=1.7,hidelinks]{hyperref}

\usepackage{orcidlink}

\begin{document}
\raggedbottom

\title{Efficient Quantization-Aware Distillation with Cross-Modal Alignment for Edge Vision–Language Models}

\titlerunning{Quantization-Aware Distillation for Edge VLMs}

\author{Jinwoo Jeon\inst{1} \and
GyuYeop Do\inst{2} \and
Yubin Lim\inst{2} \and
Nam-Joon Kim\inst{2} \and
Hyun Gon Ryu\inst{2} \and
Hyuk-Jae Lee\inst{2} \and
Byung-Jun Lee\inst{1}}

\authorrunning{J.~Jeon et al.}

\institute{Korea University\\
\email{kevin04087@korea.ac.kr} \and
Seoul National University}

\maketitle

\begin{abstract}
Large-scale vision–language models (VLMs) such as CLIP enable strong open-vocabulary reasoning, yet deploying these capabilities on resource-constrained edge devices remains challenging.
EdgeVL addresses this problem by distilling CLIP representations into lightweight multimodal encoders and applying quantization-aware training (QAT) for open-vocabulary classification (OVC) on edge devices.
However, its two-stage optimization applies different objectives for distillation and QAT, and contrastive learning is performed within the quantized student space, which can result in inconsistent optimization and reduced training efficiency. 
Moreover, identical supervision across RGB and non-RGB modalities may lead to modality imbalance. We propose a unified framework for quantized semantic distillation tailored to edge deployment. 
By jointly optimizing distillation and quantization within a unified teacher-anchored framework, our method ensures consistent training under quantization, suppressing hard negatives and enlarging decision margins.
Additionally, we design a lightweight cross-attention adapter that enhances non-RGB representations through RGB-guided semantic transfer, narrowing the modality gap. 
Experiments on EuroSAT and ScanNet demonstrate improvements in non-RGB accuracy while reducing training time.
  \keywords{Edge Vision-Language Model \and Distillation \and Open Vocabulary Classification}
\end{abstract}

\section{Introduction}
\label{sec:intro}

Recent advancements in large-scale vision–language models (VLMs), exemplified by CLIP \cite{radford2021clip}, have significantly improved the ability to connect visual perception with semantic reasoning. While these models exhibit impressive zero-shot generalization and open-set recognition capabilities, transferring their knowledge to lightweight architectures is crucial for practical deployment on resource-limited edge platforms.

\begin{figure}[t]
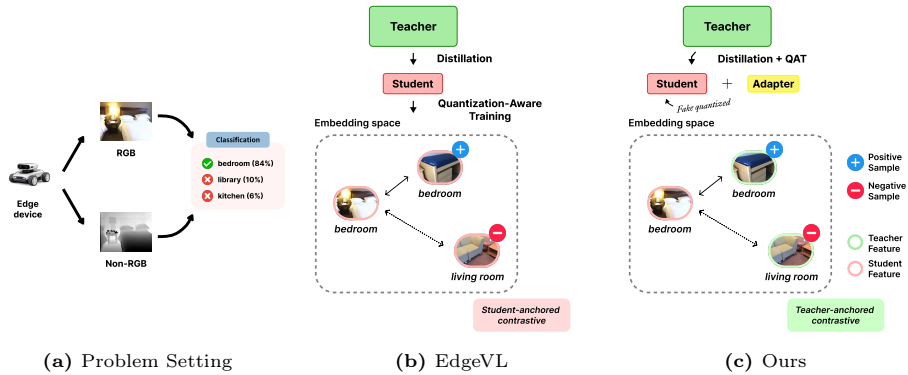

    \centering
    \begin{subfigure}[t]{0.32\linewidth}
        \centering
        \raisebox{10mm}{\includegraphics[width=\linewidth]{figure/1_a.pdf}}
        \caption{Problem Setting}
        \label{fig:fig1a}
    \end{subfigure}
    \hfill
    \begin{subfigure}[t]{0.32\linewidth}
        \centering
        \includegraphics[width=\linewidth]{figure/1_b.pdf}
        \caption{EdgeVL}
        \label{fig:fig1b}
    \end{subfigure}
    \hfill
    \begin{subfigure}[t]{0.32\linewidth}
        \centering
        \includegraphics[width=\linewidth,height=5cm,keepaspectratio]{figure/1_c.pdf}

        \caption{Ours}
        \label{fig:fig1c}
    \end{subfigure}

\caption{\textbf{Comparison of edge-oriented VLM adaptation frameworks.}
(a) An edge device collects RGB and non-RGB sensory inputs from its environment and performs open-vocabulary classification. Our goal is to adapt the edge model by distilling knowledge from a large teacher model deployed on a server.
(b) Existing approaches perform knowledge distillation and quantization-aware training (QAT) in separate stages with different objectives. During QAT, contrastive learning is applied by optimizing the distance between positive and negative samples in the quantized student embedding space.
(c) Our method unifies distillation and quantization in a single step with teacher-anchored contrastive supervision. Additionally, we introduce a lightweight cross-modal adapter that enhances non-RGB representations using RGB features.}
\label{fig:figure1}
\end{figure}

There has been growing interest in extending the capabilities of large-scale vision–language models to edge devices with limited resources. Prior work primarily focuses on improving efficiency through model compression techniques such as quantization and pruning \cite{jacob2018quantization, liu2023pdquant, han2015deep}. While these approaches reduce computational and memory overhead, they do not address challenges posed by domain shift during real-world deployment \cite{taori2020measuring, koh2021wilds,zhou2022domain}. Since human annotation is often unavailable in on-device settings, label-free adaptation is crucial for maintaining reliable performance. Furthermore, most VLMs are trained on large-scale RGB datasets, whereas edge devices frequently capture heterogeneous inputs such as depth. Enabling RGB-trained models to generalize to unseen modalities under strict computational constraints remains an open challenge. 
To address this practical setting, recent work has begun to explicitly consider edge-oriented vision–language model adaptation (\cref{fig:fig1a}). As shown in \cref{fig:fig1b}, EdgeVL \cite{cai2024self} proposes to distill the semantic alignment learned by CLIP into a lightweight vision encoder that supports both RGB and non-RGB inputs while remaining deployment-ready. EdgeVL adopts a two-stage training paradigm: it first transfers semantic knowledge through distillation, where the same supervision signal is applied uniformly across modalities, and then performs quantization-aware training (QAT) with contrastive objectives. This shared cross-modal supervision strategy enables open-vocabulary classification on heterogeneous edge sensors under limited computational budgets.

%%%%%%
% \textbf{Comparison of edge-oriented VLM adaptation frameworks.}
% (a) Edge device collects RGB and non-RGB inputs from its environment. Using those data, We aim to adapt the device's model by distilling knowlege from teacher model in larger server. In this example edge device inferences classification logits for open vocabulary set based on the inputs.
% (b) Existing apporaches, for example, EdgeVL perform distillation and QAT in two separate stages with different objectives. In QAT, contrasive learning is used where sample is trained based on the distance to positive sample and negetive sample feature in clip embedding space. All features are from edge device's inference.
% (c) 
% Our method unifies distillation and quantization with teacher-anchored contrastive supervision and introduces a lightweight cross-modal adapter to enhance non-RGB representations via RGB-guided semantic transfer. 

Despite its effectiveness, EdgeVL has a fundamental limitation: the contrastive objective is applied solely within the student’s quantized embedding space. As optimization is driven by the student’s own representations, supervision becomes self-referential, refining internal feature relations without explicitly preserving alignment to the original CLIP semantic space. Under quantization noise, this self-driven training can gradually distort semantic structure, leading to insufficient suppression of semantically similar but incorrect classes.

To address this issue, we reformulate quantization from the perspective of semantic preservation. Instead of optimizing the student within its own embedding space, we anchor supervision directly to the frozen CLIP representations throughout quantization-aware training (QAT). By introducing a teacher-anchored contrastive objective during QAT, we guide the student representations toward the CLIP semantic space. This supervision suppresses hard negatives and improves class separation under quantization while enabling efficient quantized training.

Cross-modal interaction between RGB and non-RGB modalities is an important direction in multimodal perception, yet it has not been explicitly explored in the EdgeVL setting. EdgeVL employs a shared backbone with identical supervision for RGB and non-RGB modalities, which improves parameter efficiency for edge deployment but treats each modality independently. In practice, RGB features are typically better aligned with CLIP semantics, and applying the same objective to both modalities can widen the modality gap and limit non-RGB performance. To address this limitation, we introduce a lightweight cross-modal adapter that enhances non-RGB representations through RGB guidance. The adapter is implemented using a cross-attention mechanism, where RGB features serve as keys and values and non-RGB features act as queries. This design selectively transfers semantically informative signals from RGB to non-RGB features, narrowing the modality gap while maintaining the efficiency required for edge-level deployment.
In summary, our contributions are threefold:\\ 
- We unify distillation and quantization into a single stage with teacher-anchored contrastive supervision, which increases the decision margin while improving training efficiency, resulting in up to 5.3× faster training.\\
- We propose a lightweight cross-modal adapter that enhances non-RGB representations through RGB-guided cross-modal interaction.\\
- Our method improves both non-RGB and multimodal accuracy across the evaluated backbones on EuroSAT and ScanNet while reducing training time.

\section{Related Work}
\subsection{Open-Vocabulary Recognition}
Open-vocabulary classification aims to classify images into arbitrary categories, including those not seen during training. Vision–language models (VLMs) such as CLIP \cite{radford2021clip} and ALIGN \cite{jia2021scaling} learn aligned image and text encoders using a contrastive objective that encourages corresponding image–text pairs to have higher similarity than non-matching pairs in a shared embedding space. These models have been widely adopted for open-vocabulary visual recognition tasks. Beyond image-level classification, recent works have extended this paradigm to more complex vision tasks such as object detection and segmentation. In open-vocabulary segmentation, \cite{liang2023open} generates mask proposals and learns mask embeddings aligned with CLIP’s text embedding space for mask classification. In open-vocabulary object detection, \cite{du2024lami, huang2024open} perform region classification by aligning bounding box features with text embeddings in the shared vision–language embedding space. EdgeVL \cite{cai2024self} investigates open-vocabulary classification (OVC) in an edge deployment setting.

\subsection{CLIP Distillation}
Knowledge distillation (KD) transfers knowledge from a large teacher to a compact student model to improve efficiency while preserving performance \cite{hinton2015distilling, zagoruyko2016paying}. 
In vision–language models, CLIP distillation aims to transfer the semantic representations learned by CLIP into a lightweight student encoder \cite{radford2021clip, jia2021scaling}.
Recent works explore feature-level, logit-level, and relation-level distillation strategies to better preserve the semantic structure of the teacher representations under model compression \cite{yang2024clip, yang2025clip, gou2021knowledge}. 
These approaches enable efficient open-vocabulary recognition while reducing computational overhead.

\subsection{Model Compression}
Model compression techniques reduce model size and computational cost for deployment on resource-constrained devices. 
Common strategies include pruning and quantization \cite{han2015deep, jacob2018quantization}. 
Pruning removes redundant parameters or connections in a network, reducing both memory footprint and computation while preserving predictive performance \cite{han2015deep}. 
Quantization reduces numerical precision of weights and activations to improve hardware efficiency and accelerate inference \cite{jacob2018quantization, banner2019post}. 
Quantization methods are broadly categorized into post-training quantization (PTQ) and quantization-aware training (QAT).  PTQ converts a pretrained full-precision model to low-bit representations without retraining, making it efficient for deployment but often susceptible to accuracy degradation under aggressive quantization \cite{banner2019post}. In contrast, QAT incorporates quantization effects during training so that the model learns to compensate for quantization errors, typically achieving higher accuracy under low-bit settings \cite{jacob2018quantization, esser2019learned}.

\begin{figure}[t]
    \centering
    \includegraphics[width=\linewidth]{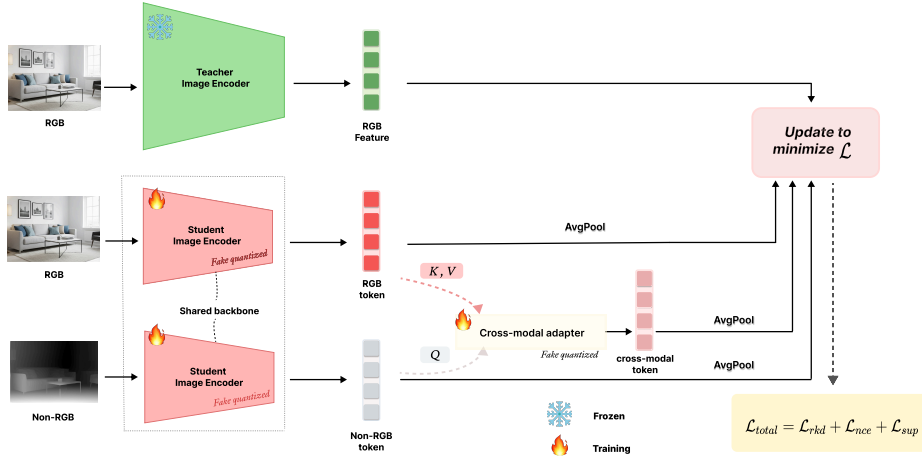}
    \caption{\textbf{Overview of our method.} A frozen CLIP encoder provides semantic supervision for a quantized student with a shared backbone supporting RGB and non-RGB inputs. RGB features are directly distilled toward the teacher embedding, while non-RGB features are enhanced through a cross-modal attention module using non-RGB queries and RGB keys and values. The student is trained with teacher-anchored contrastive learning and an RKD loss to preserve geometric structure in the embedding space. The student encoder and adapter are jointly optimized with quantization-aware training for efficient edge deployment, while the teacher remains frozen.}
    \label{fig:figure2}
\end{figure}
\section{Method}
\subsection{Problem Formulation}

We study annotation-free adaptation of an open-vocabulary vision model to heterogeneous edge sensors. An overview of the proposed framework is shown in \cref{fig:figure2}.
In practical deployments, multiple co-located sensors capture synchronized observations of the same scene, while semantic labels are unavailable.

\paragraph{Unlabeled Synchronized Multi-Modal Data and Training.}
Let $\mathcal{U}=\{(\mathbf{x}_i^{r},\mathbf{x}_i^{r'})\}_{i=1}^{M}$ denote an unlabeled adaptation set of synchronized sensor pairs, 
where $\mathbf{x}_i^{r}$ is an RGB observation and $\mathbf{x}_i^{r'}$ is a corresponding heterogeneous observation (e.g., depth or thermal) captured from the same scene. 
We assume access to a frozen teacher VLM image encoder $E_0$ (e.g., CLIP), which defines a semantic embedding space in $\mathbb{R}^{D}$. 
Our objective is to learn an edge-efficient student image encoder $E_\theta$ that supports heterogeneous modalities while remaining aligned with the teacher semantic space. 
During training, we leverage pseudo-labels predicted by the teacher model such as CLIP to distill semantic knowledge into the student encoder. 
The goal is to preserve the teacher's open-vocabulary capabilities while adapting the student to heterogeneous inputs and enabling transfer across datasets.

\paragraph{Open-Vocabulary Inference.}
Given a class vocabulary $\mathcal{C}$ and the frozen text encoder $T$ from the teacher VLM, 
we compute normalized text embeddings for all $c \in \mathcal{C}$. 
For single-modality inference with a test input $\mathbf{x}_i^{m}$, where $m \in \{r, r'\}$, prediction is performed via cosine similarity:
\[
\hat{y}_i^{m}
=
\arg\max_{c\in\mathcal{C}}
\left\langle
\frac{E_\theta(\mathbf{x}_i^{m})}{\|E_\theta(\mathbf{x}_i^{m})\|_2},
\frac{T(c)}{\|T(c)\|_2}
\right\rangle .
\]

\subsection{End-to-End Quantization-Aware Distillation}
\begin{figure}[t]
\centering

\begin{subfigure}[t]{0.32\linewidth}
    \centering
    \raisebox{0.5cm}{\includegraphics[width=\linewidth]{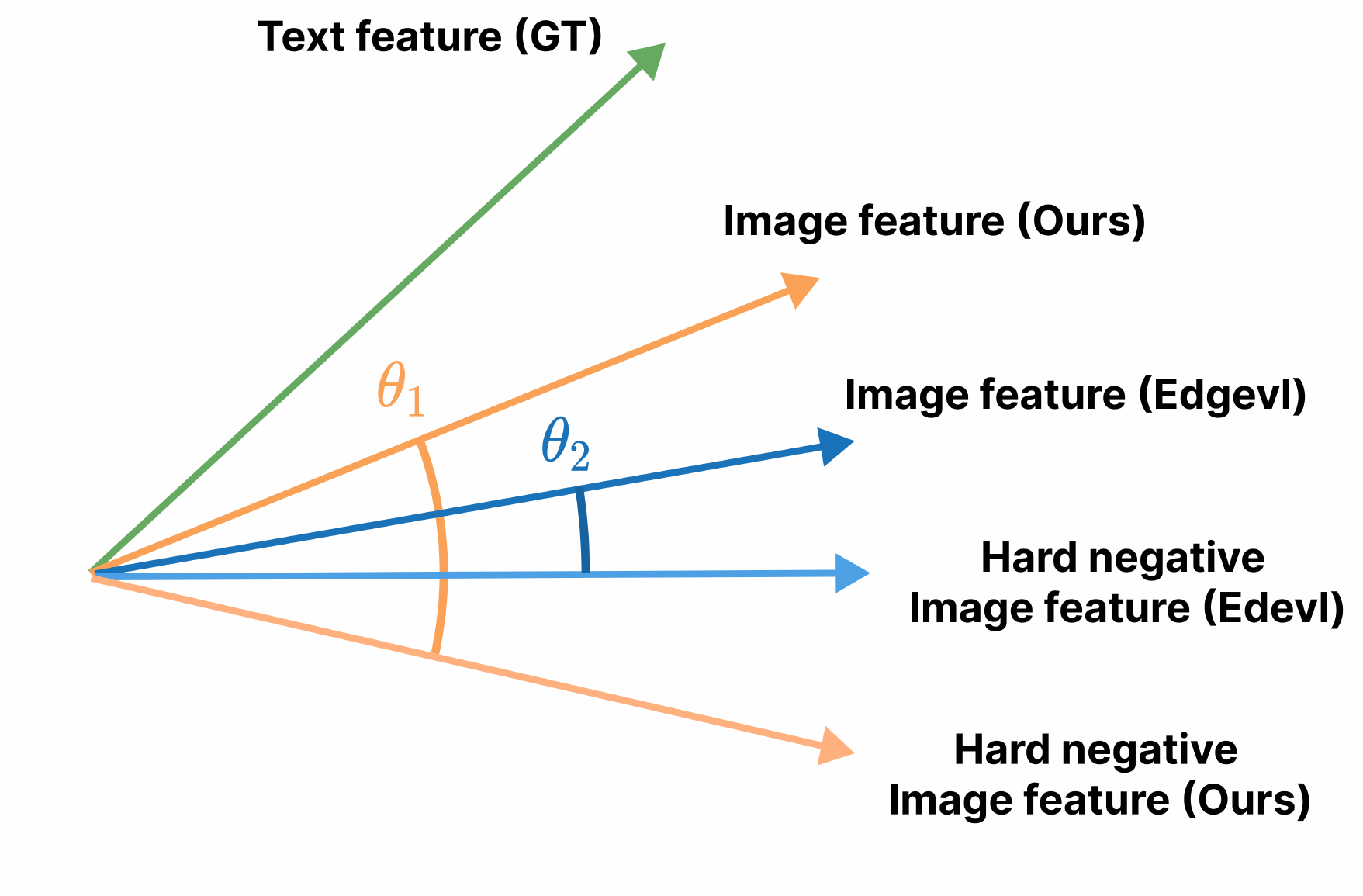}}
    \caption{margin angle}
    \label{fig:fig3a}
\end{subfigure}
\hfill
\begin{subfigure}[t]{0.32\linewidth}
    \centering
    \includegraphics[width=\linewidth]{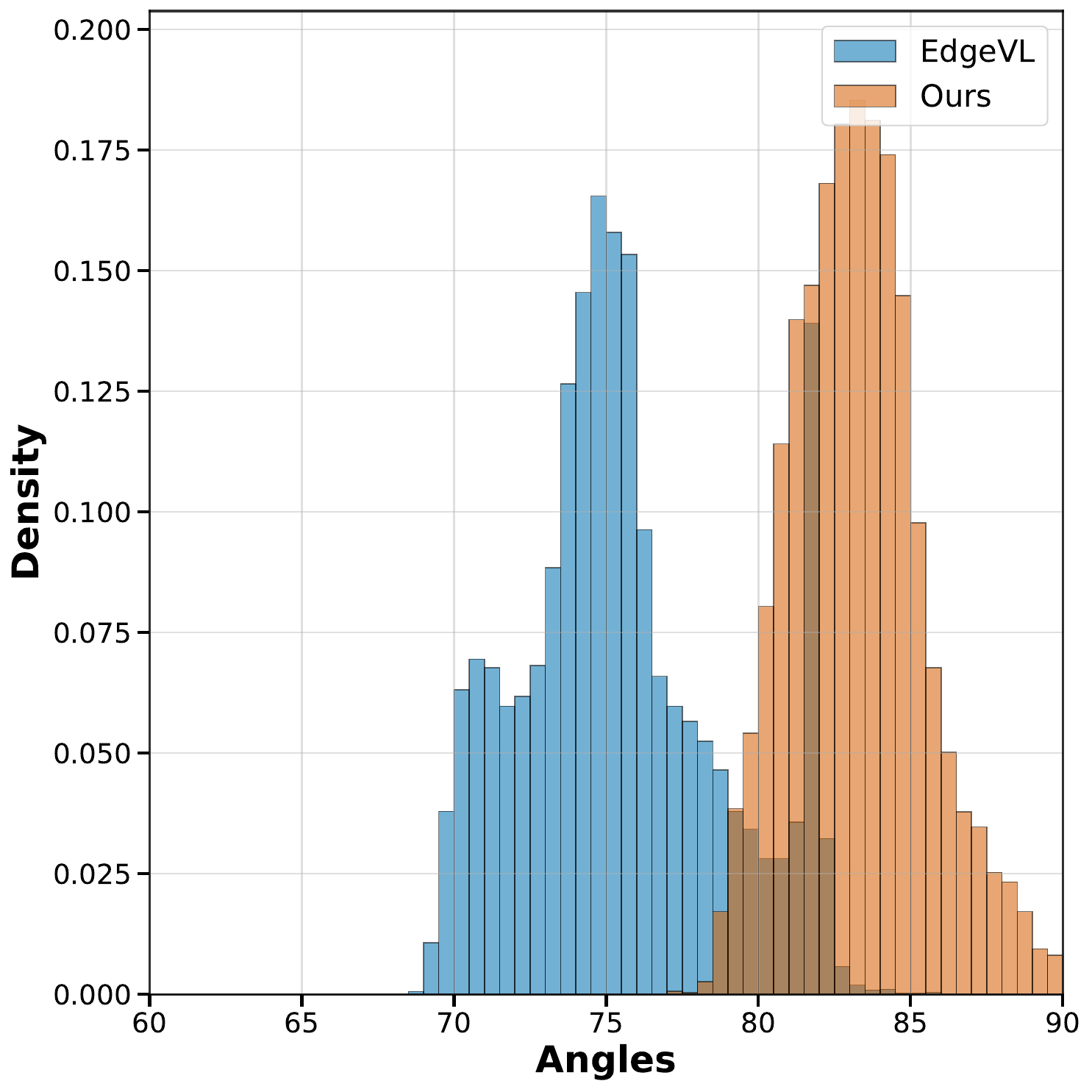}
    \caption{hard negative angle}
    \label{fig:fig3b}
\end{subfigure}
\hfill
\begin{subfigure}[t]{0.32\linewidth}
    \centering
    \includegraphics[width=\linewidth]{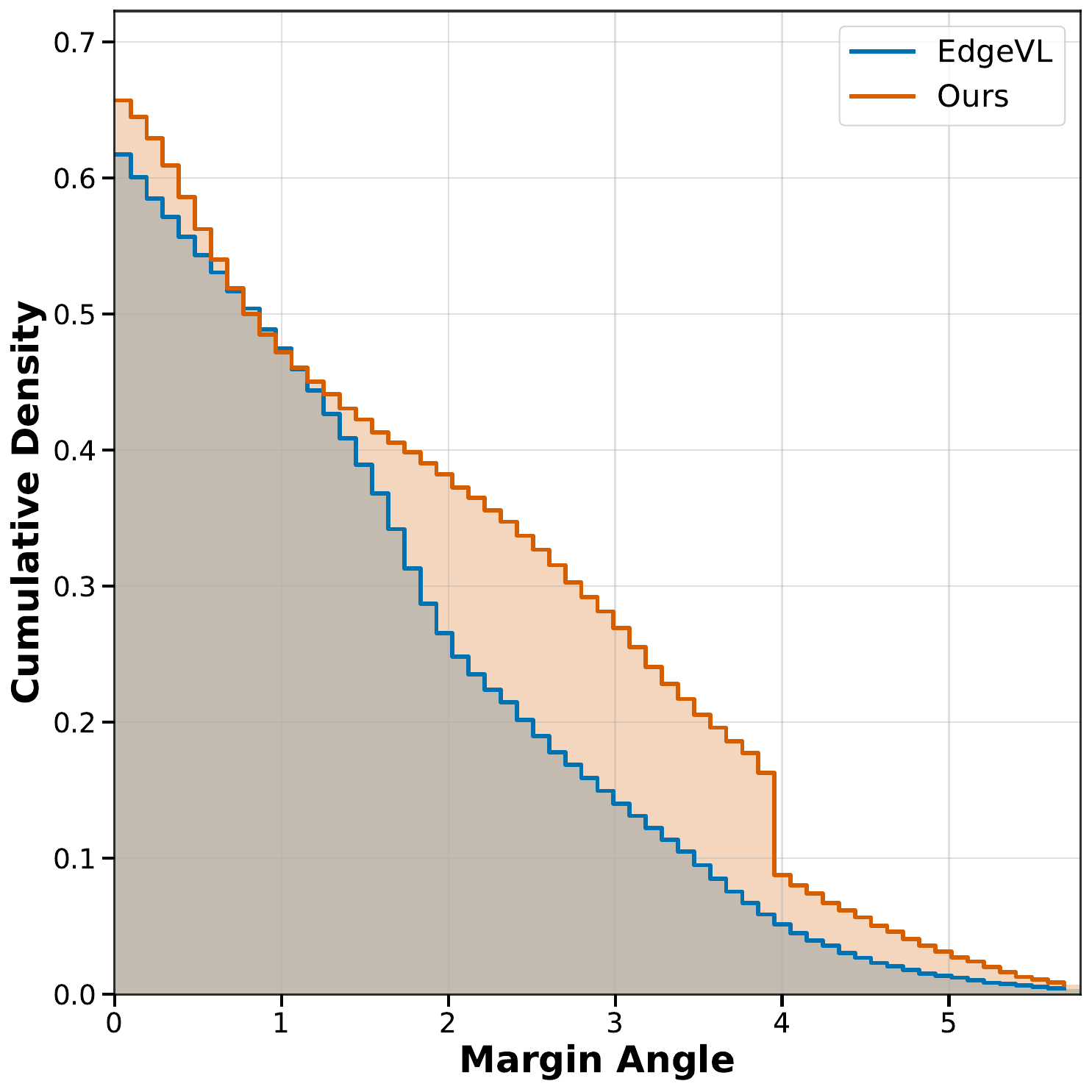}
    \caption{cumulative density}
    \label{fig:fig3c}
\end{subfigure}

\caption{\textbf{Decision margin visualization.}
(a) Decision margin angle between the image embedding and the hardest negative class.
Our method preserves alignment with the ground-truth text embedding while increasing
angular separation from hard negatives.
(b) Compared to EdgeVL, our model yields a larger angle between the ground-truth text
embedding and the hardest negative class, resulting in a wider decision margin.
(c) Cumulative density whose margin angle exceeds a given threshold.
Our method retains more samples at larger margins, indicating improved separation
from hard negatives.}
\label{fig:fig3}
\end{figure}

EdgeVL~\cite{cai2024self} adopts a two-stage pipeline consisting of full-precision distillation followed by QAT with a triplet-based objective. However, the triplet loss is optimized solely within the quantized student embedding space using self-mined positives and negatives, which can deviate from the global CLIP geometry under quantization noise. As a result, semantically similar but incorrect classes may remain insufficiently separated. In contrast, we integrate contrastive supervision directly into the distillation stage and anchor student representations to the frozen teacher space. By aligning student embeddings with the original CLIP representations, our teacher-anchored objective preserves global semantic consistency and suppresses hard negative confusion under quantization (see \cref{fig:fig3}).

\paragraph{Relational Knowledge Distillation (RKD).}
QAT can distort the geometry of the embedding space due to quantization noise.
To preserve the global relational structure of the teacher representations under
such perturbations, we first employ relational knowledge distillation
(RKD)~\cite{park2019relational}, which transfers pairwise and angular relations
between samples from the teacher to the student.
Given a mini-batch $\{\mathbf{x}_i\}_{i=1}^{B}$ (where each $\mathbf{x}_i$ can
be either $\mathbf{x}_i^{r}$ or $\mathbf{x}_i^{r'}$ depending on the training
stream), we define normalized student and teacher embeddings:
\[
\mathbf{s}_i = \frac{E_\theta(\mathbf{x}_i)}{\|E_\theta(\mathbf{x}_i)\|_2},
\qquad
\mathbf{t}_i = \frac{E_0(\mathbf{x}_i)}{\|E_0(\mathbf{x}_i)\|_2}.
\]
Following \cite{park2019relational}, RKD aligns distance- and angle-wise
relations:
\[
\mathcal{L}_{\mathrm{rkd\text{-}D}}
=
\sum_{i<j}
l_\delta\!\left(
\|\mathbf{s}_i-\mathbf{s}_j\|_2,\
\|\mathbf{t}_i-\mathbf{t}_j\|_2
\right),
\]
\[
\mathcal{L}_{\mathrm{rkd\text{-}A}}
=
\sum_{i<j<k}
l_\delta\!\left(
\cos\angle(\mathbf{s}_i,\mathbf{s}_j,\mathbf{s}_k),\
\cos\angle(\mathbf{t}_i,\mathbf{t}_j,\mathbf{t}_k)
\right),
\]
where
\[
\cos\angle(\mathbf{s}_i,\mathbf{s}_j,\mathbf{s}_k)
=
\left\langle\,
\frac{\mathbf{s}_j - \mathbf{s}_i}{\|\mathbf{s}_j - \mathbf{s}_i\|_2},\
\frac{\mathbf{s}_k - \mathbf{s}_i}{\|\mathbf{s}_k - \mathbf{s}_i\|_2}
\,\right\rangle
\]
denotes the cosine of the angle at vertex $\mathbf{s}_i$ formed by the triplet
$(\mathbf{s}_i, \mathbf{s}_j, \mathbf{s}_k)$, and analogously for
$\cos\angle(\mathbf{t}_i,\mathbf{t}_j,\mathbf{t}_k)$, and $l_\delta$ denotes
the Huber loss~\cite{huber1992robust}. RKD is defined as follows:
\[
\mathcal{L}_{\mathrm{rkd}} = \mathcal{L}_{\mathrm{rkd\text{-}D}} +
\lambda\,\mathcal{L}_{\mathrm{rkd\text{-}A}}.
\]

\paragraph{Teacher-Anchored Contrastive Objectives.}
While RKD helps preserve the relational geometry of the teacher space,
it does not explicitly enforce \emph{instance-to-teacher} consistency.
To stabilize semantic grounding under quantization, we incorporate
teacher-anchored contrastive supervision that directly ties each quantized
student embedding to its corresponding frozen teacher embedding. Specifically, we adopt a symmetric InfoNCE objective~\cite{radford2021clip}
between normalized student and teacher embeddings:
\[
\mathcal{L}_{\mathrm{nce}}
=
-\frac{1}{2B}\sum_{i=1}^{B}
\left[
\log \frac{\exp(\mathbf{s}_i^\top \mathbf{t}_i)}
          {\sum_{j=1}^{B} \exp(\mathbf{s}_i^\top \mathbf{t}_j)}
+
\log \frac{\exp(\mathbf{t}_i^\top \mathbf{s}_i)}
          {\sum_{j=1}^{B} \exp(\mathbf{t}_i^\top \mathbf{s}_j)}
\right].
\]
In contrast to self-referential contrastive learning performed purely within
the student space~\cite{cai2024self}, our formulation uses frozen teacher
embeddings as immutable anchors, thereby explicitly suppressing representation
drift caused by quantization noise.
Low-confidence teacher predictions can introduce noisy pseudo-labels into
class-aware supervision. We therefore retain training samples whose teacher
confidence exceeds a threshold $\tau$ to reduce ambiguous pseudo-labels.

The instance-wise objective treats other samples as negatives even when they
share the same pseudo-label. It therefore does not explicitly preserve
category-level structure and can push same-class samples apart in the student
space. To inject structured semantic knowledge from the teacher, we further
incorporate a supervised contrastive objective built upon CLIP-derived
pseudo-labels $\tilde{y}_i$. Define the positive set
\[
\mathcal{P}(i)=\{\,j\neq i \mid \tilde{y}_j=\tilde{y}_i\,\}.
\]
We minimize
\[
\mathcal{L}_{\mathrm{sup}}
=
-\frac{1}{B}\sum_{i=1}^{B}
\frac{1}{|\mathcal{P}(i)|}
\sum_{j\in\mathcal{P}(i)}
\log
\frac{\exp(\mathbf{s}_i^\top \mathbf{t}_j)}
     {\sum_{k:\,k\notin\mathcal{P}(i)} \exp(\mathbf{s}_i^\top \mathbf{t}_k)}.
\]
This class-aware term promotes intra-class compactness and enlarges inter-class
angular margins \emph{with respect to teacher semantics}, complementing the
instance-level anchoring effect of $\mathcal{L}_{\mathrm{nce}}$.

\paragraph{Distillation Objective.}
We train the student encoder by minimizing the combined teacher-anchored
objective evaluated on its normalized embeddings $\mathbf{s}$:
\[
\mathcal{L}_{\mathrm{Distill}}(\mathbf{s})
=
\mathcal{L}_{\mathrm{rkd}}(\mathbf{s})
+
\mathcal{L}_{\mathrm{nce}}(\mathbf{s})
+
\mathcal{L}_{\mathrm{sup}}(\mathbf{s}).
\]

\paragraph{Cross-Modal Attention Adapter.}
EdgeVL adopts a shared backbone with identical supervision for RGB and non-RGB modalities, but does not explicitly model cross-modal interaction. Since RGB features are typically better aligned with the teacher semantic space, applying the same supervision to both modalities can bias optimization toward RGB-dominant representations, limiting the semantic quality of non-RGB features. To address this limitation, we introduce a lightweight cross-modal adapter based on multi-head attention \cite{vaswani2017attention}. The adapter enhances non-RGB representations by allowing non-RGB tokens to attend to semantically richer RGB tokens.
Given paired RGB and non-RGB inputs $(\mathbf{x}_i^{r}, \mathbf{x}_i^{r'})$, the student backbone produces token embeddings
\[
\mathbf{Z}_i^{r}, \mathbf{Z}_i^{r'} \in \mathbb{R}^{L \times D},
\]
where $L$ denotes the number of output tokens and $D$ is the token embedding dimension. Modality-specific pooled embeddings are obtained as
\[
\mathbf{s}_i^{r}
=
\frac{\mathrm{Pool}(\mathbf{Z}_i^{r})}{\|\mathrm{Pool}(\mathbf{Z}_i^{r})\|_2},
\qquad
\mathbf{s}_i^{r'}
=
\frac{\mathrm{Pool}(\mathbf{Z}_i^{r'})}{\|\mathrm{Pool}(\mathbf{Z}_i^{r'})\|_2}.
\]
To enhance the non-RGB representation, we apply cross-attention using non-RGB tokens as queries and RGB tokens as keys and values. For simplicity, we omit the sample index $i$ below:
\[
\mathbf{Q} = \mathbf{Z}^{r'}\mathbf{W}_Q, \quad
\mathbf{K} = \mathbf{Z}^{r}\mathbf{W}_K, \quad
\mathbf{V} = \mathbf{Z}^{r}\mathbf{W}_V,
\]
where $\mathbf{W}_Q,\mathbf{W}_K,\mathbf{W}_V,\mathbf{W}_O \in \mathbb{R}^{D \times D}$ are learnable projection matrices. The enhanced non-RGB tokens are computed as
\[
\mathbf{A}_h
=
\mathrm{softmax}\!\left(
\frac{\mathbf{Q}_h \mathbf{K}_h^\top}{\sqrt{D_h}}
\right)\mathbf{V}_h .
\]
\[
\mathbf{U}^{r'}
=
\mathrm{Concat}(\mathbf{A}_1,\dots,\mathbf{A}_H)\mathbf{W}_O ,
\]
where $H$ is the number of attention heads and $D_h = D/H$ is the dimension of each head.
The residual connection defines the adapted tokens as
\[
\tilde{\mathbf{Z}}_i^{r'} = \mathbf{Z}_i^{r'} + \mathbf{U}_i^{r'}.
\]
The final enhanced non-RGB embedding is then obtained by pooling these adapted tokens:

\[
\tilde{\mathbf{s}}_i^{r'}
=
\frac{\mathrm{Pool}(\tilde{\mathbf{Z}}_i^{r'})}{\|\mathrm{Pool}(\tilde{\mathbf{Z}}_i^{r'})\|_2}.
\]
To preserve efficiency for edge deployment, the adapter uses only a single cross-attention layer with a residual connection.

\input{table/main_table} 

\paragraph{Final Training Objective with the Adapter.}
We apply the same teacher-anchored distillation objective to the normalized embeddings of RGB, non-RGB, and cross-modal features.
\[
\mathcal{L}_{\mathrm{total}}
=
\mathcal{L}_{\mathrm{Distill}}(\mathbf{s}^{r})
+
\mathcal{L}_{\mathrm{Distill}}(\mathbf{s}^{r'})
+
\mathcal{L}_{\mathrm{Distill}}(\tilde{\mathbf{s}}^{r'}).
\]
\section{Experiments}
\subsection{Settings}
\textbf{Implementation Details} We adopt the CLIP ViT-g-14 (ViT-G) model released by OpenCLIP~\cite{ilharco_gabriel_2021_5143773} as the teacher network. For the student architecture, we employ ViT-S~\cite{li2022qvit}, DAT-T~\cite{han2022data}, and Swin-T~\cite{liu2021swin} as shared backbone networks, following the setting of~\cite{cai2024self}. During training, the student models are optimized using AdamW~\cite{loshchilov2017fixing} with an initial learning rate of $2\times10^{-4}$ and a weight decay of 0.05. A cosine annealing schedule gradually decreases the learning rate to $5\times10^{-6}$ over 30 epochs. For the confidence threshold $\tau$, we empirically set it to 0.25 to balance data utilization and label noise, following~\cite{cai2024self}. For the CLIP text encoder, we use prompt templates of the form ``a photo of a \{scene category\}.'' or ``a satellite image of a \{scene category\}.'' to generate text embeddings. To enable cross-modal interaction, we introduce a lightweight cross-modal attention adapter implemented as a single-layer multi-head attention module. Although this module enables explicit fusion between RGB and non-RGB features, it adds 9~MB after quantization, equivalent to approximately 16\% of the reported 56~MB Swin-T backbone size. \\
\textbf{Dataset} We conduct experiments on two benchmarks, EuroSAT~\cite{helber2018eurosat} and ScanNet~\cite{dai2017scannet}. Following \cite{cai2024self}, we apply a subsampling factor of 100 on ScanNet to reduce redundancy, resulting in 18,900 training, 5,300 validation, and 2,100 test RGB-D images across 21 indoor scene categories. Since the official test split does not provide labels, evaluation is conducted on the validation set. EuroSAT contains 27,000 satellite images across 13 spectral bands and 10 land-use classes, which we randomly split into 13,500 training and 13,500 testing samples. \\
\textbf{Baselines} We adapt several related methods for comparison, including EdgeVL \cite{cai2024self} as a strong baseline. CMKD \cite{hong2022crossmodality}, Fida~\cite{thoker2019crossmodal}, and CQD \cite{su2017adapting} are modified to align non-RGB representations with those of a pre-trained RGB model and the CLIP visual encoder. SKD \cite{yang2022mixskd} is adapted for joint RGB and non-RGB training via hybrid mixup, while approaches inspired by Frank and Gupta \cite{hafner2022crossmodal, hoffman2016crossmodal} are considered for cross-modal transfer and unified embedding alignment. Following the original EdgeVL setting, we evaluate both EdgeVL and our method under 8-bit quantization for fair comparison. \\
\input{table/multimodal_accuracy}
\input{table/latency} \textbf{Evaluation Metric} We use top-1 accuracy as the evaluation metric.
For both RGB and non-RGB evaluation, performance is measured using the modality-specific features extracted from the shared backbone. The reported Avg score denotes the arithmetic mean of the RGB and non-RGB accuracies. In addition, we report multimodal accuracy to assess the effectiveness of feature fusion when both RGB and non-RGB inputs are available. This metric reflects performance under joint modality inference. For our method, multimodal performance is computed from the fused representation obtained via cross-attention, where non-RGB features serve as queries and RGB features act as keys and values. In contrast, since EdgeVL does not employ an explicit cross-attention mechanism, its multimodal representation is formed by simply averaging the RGB and non-RGB features extracted from the shared backbone. All other baselines are reported in full precision (FP32) using their best-performing backbones. In the main results, we report performance under the configuration that achieves the best non-RGB performance. For the cross-modal setting, results are reported using the configuration that yields the highest cross-modal performance.\\
\input{table/cross_dataset}
\input{table/crossdata_multimodal_accuracy}
\subsection{Results}
\textbf{Main Result} As shown in \cref{tab:open-vocabulary-classification-overall}, under the same shared-backbone setting, our model improves non-RGB accuracy across all reported backbones on both datasets. RGB accuracy also improves in most cases, with a small decrease for Swin-T on EuroSAT (67.1\% to 66.7\%). On EuroSAT, we observe up to a 4.7 percentage point gain in non-RGB performance, while on ScanNet the improvement reaches 4.9 percentage points. In \cref{tab:multimodal_accuracy_scannet}, we further compare multimodal accuracy on EuroSAT and ScanNet. The results show consistent improvements across datasets and backbone architectures, indicating that the proposed cross-modal adapter effectively facilitates cross-modal alignment. \\
\textbf{Efficiency} To evaluate efficiency, we measure both training time and inference throughput. For inference efficiency, throughput is measured after converting the trained model from ONNX to TensorRT.  As shown in \cref{tab:training_time_comparison}, our method significantly improves training efficiency, reducing the total training time by 81\%, corresponding to a 5.3× speedup compared to the baseline.  For inference efficiency, we additionally measure throughput when the cross-modal attention module is enabled. Although the proposed adapter introduces additional computation, throughput decreases from 461 to 351 images/s for ViT-S (23.9\%) and from 415 to 359 images/s for Swin-T (13.5\%). These results quantify the inference-throughput trade-off associated with the proposed method.

\subsection{Cross-Dataset Generalization}
To evaluate cross-dataset generalization, we train the model on ScanNet and test on NYUv2 \cite{silberman2012nyuv} and SUN-RGBD \cite{song2015sun}. SUN-RGBD contains 5,285 training and 5,050 testing RGB-D images annotated with 19 scene categories, collected from multiple RGB-D cameras. NYUv2 consists of 795 training and 654 testing images across 10 scene classes captured using a Kinect sensor. As shown in \cref{tab:cross-dataset}, our method consistently improves performance for both ViT-S and DAT-T backbones, indicating strong cross-dataset generalization. A similar trend is observed in multimodal accuracy for these two backbones (\cref{tab:multimodal_accuracy}). However, Swin-T shows lower non-RGB and RGB accuracy on SUN-RGBD and lower multimodal accuracy on both unseen datasets than EdgeVL, indicating that the generalization gains are backbone-dependent.

\subsection{Ablation Study}
\textbf{Ablation on Loss Components} We conduct ablation studies on ScanNet \cite{dai2017scannet} using a shared backbone ViT-S \cite{cai2022efficientvit} to analyze the contribution of each loss component. Starting from a baseline loss, we progressively add additional objectives and evaluate the resulting classification accuracy. As shown in \cref{tab:loss_ablation_style}, using only RKD during QAT leads to poor performance, suggesting that preserving relational structure alone is insufficient to maintain semantic alignment under quantization. Adding InfoNCE significantly improves performance by enforcing instance-level alignment and enlarging the margin between positive and hard negative samples. However, instance-level contrastive learning alone is insufficient for complex indoor scenes such as ScanNet. Adding the pseudo-label-based supervised contrastive term yields the highest RGB and non-RGB accuracy among the evaluated loss configurations.
\input{table/loss_ablation}
\input{table/ablation}
\par
\textbf{Robustness under Low-bit Quantization.}
To evaluate robustness under more aggressive quantization, we further reduce the precision to 6-bit and 4-bit on EuroSAT. Performance remains nearly unchanged at 6-bit, indicating robustness to moderate quantization. At 4-bit, accuracy decreases due to the limited numerical resolution (16 quantization levels), which can introduce larger distortions in cosine similarity computations. Despite this, the model still maintains over 50\% non-RGB accuracy, demonstrating reasonable performance even under aggressive quantization.\par
\textbf{Impact of the Cross-Modal Adapter} To evaluate the effectiveness of the proposed cross-modal adapter, we remove both the adapter module and the cross-modal supervision, leaving only the shared backbone. In this configuration, RGB and non-RGB inputs are processed independently without explicit cross-modal interaction. The multimodal prediction is obtained by averaging the RGB and non-RGB feature representations. On EuroSAT, the reported accuracy gains over the configuration without the adapter range from 1.3 to 2.2 percentage points (\cref{tab:adapter_ablation}). Because this ablation removes both the adapter and cross-modal supervision, it measures their combined contribution rather than isolating the effect of the adapter architecture.\par
\section{Conclusion}
We propose a unified quantization-aware distillation framework for edge VLMs that preserves CLIP semantics under low-bit training via teacher-anchored contrastive supervision. By anchoring quantized student embeddings to the frozen teacher space, the method suppresses hard negatives and enlarges angular margins, mitigating semantic drift caused by quantization. We further introduce a lightweight cross-modal attention adapter that transfers semantic cues from RGB to non-RGB features in shared-backbone settings. Experiments on EuroSAT and ScanNet demonstrate improvements in non-RGB and multimodal accuracy across the evaluated backbones.
% \clearpage\mbox{}Page \thepage\ of the manuscript.
% \clearpage\mbox{}Page \thepage\ of the manuscript.
% \clearpage\mbox{}Page \thepage\ of the manuscript.
% \clearpage\mbox{}Page \thepage\ of the manuscript.
% \clearpage\mbox{}Page \thepage\ of the manuscript. This is the last page.
\par\vfill\par
\clearpage  % TODO FINAL: This \clearpage needs to be removed from both review and camera-ready versions.

\bibliographystyle{splncs04}
\bibliography{main}
\end{document}

%% file: table/main_table.tex
\begin{table}[t]
  \centering
  \renewcommand\arraystretch{1.2}
  \setlength\tabcolsep{4pt}  
  \caption{Overall accuracy comparison. \nrgb\ and \rgb\ indicate the top-1 accuracy for non-RGB and RGB inputs, respectively, and \avg\ denotes their average, all reported as percentages. The same notation applies to the subsequent tables. For each backbone, results that improve over EdgeVL are highlighted in bold.}
  \label{tab:open-vocabulary-classification-overall}
  \scalebox{0.87}{
  \begin{tabular}{l|l|ccc|ccc} 
  \hline
  \noalign{\vskip 2pt}
  \multirow{2}{*}{Methods} & \multirow{2}{*}{Bits} 
  & \multicolumn{3}{c|}{ScanNet (\%) $\uparrow$ } 
  & \multicolumn{3}{c}{EuroSAT (\%) $\uparrow$} \\ 
  & & \nrgb & \rgb & \avg & \nrgb & \rgb & \avg \\ 
  \noalign{\vskip 2pt}
  \hline
  \noalign{\vskip 2pt}

  Pretrained CLIP-B \cite{radford2021clip} & F32 & 4.5 & 36.2 & 20.4 & 16.8 & 40.4 & 28.6 \\
  Pretrained CLIP-G \cite{radford2021clip} & F32 & 6.2 & 47.3 & 26.8 & 16.9 & 54.0 & 35.5 \\
  
  \noalign{\vskip 2pt}
  \hline
  \noalign{\vskip 2pt}

  Frank \cite{hafner2022crossmodal}  & F32 & 8.3 & 21.7 & 15.0 & 49.2 & 37.9 & 43.5 \\
  Gupta \cite{hoffman2016crossmodal} & F32 & 16.0 & 17.5 & 16.8 & 54.2 & 42.4 & 48.3 \\
  CMKD \cite{hong2022crossmodality} (non-RGB) & F32 & 37.8 & 11.5 & 24.6 & 61.2 & 34.4 & 47.8 \\
  CMKD \cite{hong2022crossmodality} (RGB) & F32 & 4.0 & 42.5 & 23.2 & 20.1 & 62.4 & 41.2 \\
  Fida \cite{thoker2019crossmodal} & F32 & 38.9 & 5.8 & 22.3 & 56.7 & 20.3 & 38.5 \\
  CQD \cite{su2017adapting} & F32 & 40.1 & 6.7 & 23.4 & 62.4 & 36.4 & 49.4 \\
  SKD \cite{yang2022mixskd} & F32 & 31.2 & 37.8 & 34.5 & 22.9 & 50.3 & 36.6 \\

  \noalign{\vskip 2pt}
  \hline
  \noalign{\vskip 2pt}

  EdgeVL (DAT-T) \cite{cai2024self} & Int8 & 47.9 & 52.0 & 49.9 & 61.0 & 65.7 & 63.3 \\
  EdgeVL (Swin-T) \cite{cai2024self}& Int8 & 46.0 & 48.7 & 47.4 & 61.3 & 67.1 & 64.2 \\
  EdgeVL (ViT-S) \cite{cai2024self}  & Int8 & 42.0 & 47.5 & 44.7 & 62.9 & 66.8 & 64.8 \\

  \noalign{\vskip 2pt}
  \hline
  \noalign{\vskip 2pt}

  % ----------- Ours 추가 -----------
  Ours (DAT-T)  & Int8 & \textbf{49.0} & \textbf{52.2} & \textbf{50.6} & \textbf{65.7} & \textbf{67.9} & \textbf{66.8} \\
  Ours (Swin-T)  & Int8 & \textbf{47.2} & \textbf{50.4} & \textbf{48.8} & \textbf{65.0} & 66.7 & \textbf{65.9} \\
  Ours (ViT-S)   & Int8 & \textbf{46.9} & \textbf{51.3} & \textbf{49.1} & \textbf{64.7} & \textbf{69.1} & \textbf{66.9} \\

  \noalign{\vskip 2pt}
  \hline
  \end{tabular}
  }
\end{table}

%% file: table/multimodal_accuracy.tex
\begin{table}[t]
  \centering
  \renewcommand\arraystretch{1.2}
  \setlength\tabcolsep{6pt}
  \caption{Multimodal accuracy comparison between EdgeVL and Ours.}
  \label{tab:multimodal_accuracy_scannet}
  \scalebox{0.95}{
  \begin{tabular}{l|c|cc}
  \hline
  \noalign{\vskip 2pt}
  Methods & Bits & ScanNet (\%) $\uparrow$ & EuroSAT (\%) $\uparrow$ \\
  \noalign{\vskip 2pt}
  \hline
  \noalign{\vskip 2pt}

  EdgeVL (DAT-T)  \cite{cai2024self} & Int8 & 51.1 & 65.5 \\
  EdgeVL (Swin-T) \cite{cai2024self} & Int8 & 49.4 & 66.8 \\
  EdgeVL (ViT-S)  \cite{cai2024self} & Int8 & 46.4 & 66.9 \\

  \noalign{\vskip 2pt}
  \hline
  \noalign{\vskip 2pt}

  Ours (DAT-T)  & Int8 & \textbf{51.6} & \textbf{71.4} \\
  Ours (Swin-T) & Int8 & \textbf{50.7} & \textbf{72.0} \\
  Ours (ViT-S)  & Int8 & \textbf{51.0} & \textbf{70.8} \\

  \noalign{\vskip 2pt}
  \hline
  \end{tabular}
  }
\end{table}

%% file: table/latency.tex
\begin{table}[t]
  \centering
  \renewcommand\arraystretch{1.2}
  \setlength\tabcolsep{4pt}   
  \caption{Overall efficiency comparison in terms of training time and throughput.}
  \label{tab:training_time_comparison}
  \scalebox{0.9}{
  \begin{tabular}{l|c|c|c} 
    \hline
    \noalign{\vskip 2pt}
    Methods & Bits & Training Time $\downarrow$ & Throughput $\uparrow$ \\
    \noalign{\vskip 2pt}
    \hline
    \noalign{\vskip 2pt}

    EdgeVL (ViT-S)  & Int8 & 12 hours
                    & \textbf{461 images/s} \\

    EdgeVL (Swin-T) & Int8 & 12 hours
                    & \textbf{415 images/s} \\

    \noalign{\vskip 2pt}
    \hline
    \noalign{\vskip 2pt}

    Ours (ViT-S)  & Int8 
                 & \textbf{2.25 hours} ($\downarrow 81\%$)
                 & 351 images/s ($\downarrow 23.9\%$) \\

    Ours (Swin-T) & Int8 
                 &  \textbf{2.25 hours} ($\downarrow 81\%$)
                 & 359 images/s ($\downarrow 13.5\%$) \\

    \noalign{\vskip 2pt}
    \hline
  \end{tabular}
  }
\end{table}

%% file: table/cross_dataset.tex
\label{sub:cross_dataset_performance}

\begin{table}[!t]
  \centering
  \renewcommand\arraystretch{1.2}
  \setlength\tabcolsep{4pt}  
  \caption{Accuracy on unseen datasets after training on ScanNet. For each backbone, results that improve over EdgeVL are highlighted in bold.}
  \vspace{-5pt}
  \label{tab:cross-dataset}
  \scalebox{0.85}{
  \begin{tabular}{l|l|ccl|ccl} 
  \hline
  \noalign{\vskip 2pt}
  \multirow{2}{*}{Methods} & \multirow{2}{*}{Bits} 
  & \multicolumn{3}{c|}{NYUv2 (\%)}
  & \multicolumn{3}{c}{SUN-RGBD (\%)} \\
  & & \nrgb & \rgb & \multicolumn{1}{c|}{\avg} 
    & \nrgb & \rgb & \multicolumn{1}{c}{\avg}  \\ 
  \noalign{\vskip 2pt}
  \hline
  \noalign{\vskip 2pt}

  Pre-trained CLIP-G & F32 
  & 25.7 & 69.7 & 47.7
  & 18.0 & 54.3 & 36.2 \\

  Pre-trained CLIP-B & F32 
  & 22.6 & 62.2 & 42.4 
  & 15.2 & 47.2 & 31.2 \\  

  \noalign{\vskip 2pt}
  \hline
  \noalign{\vskip 2pt}

  EdgeVL (DAT-T)  & Int8 
  & 51.1 & 54.3 & 52.7 
  & 28.6 & 31.8 & 30.2 \\

  EdgeVL (Swin-T) & Int8 
  & 43.4 & 43.3 & 43.4 
  & 30.0 & 31.4 & 30.7 \\

  EdgeVL (ViT-S)  & Int8 
  & 41.0 & 40.5 & 40.8 
  & 25.8 & 28.0 & 27.0 \\

  \noalign{\vskip 2pt}
  \hline
  \noalign{\vskip 2pt}

  % -------- Ours 추가 --------
  Ours (DAT-T)  & Int8 
  & \textbf{53.8} & \textbf{57.7} & \textbf{55.8}
  & \textbf{29.2} & \textbf{34.4} & \textbf{31.7} \\

  Ours (Swin-T) & Int8 
  & \textbf{47.9} & 43.1 & \textbf{45.5}
  & 24.8 & 30.0 & 27.4 \\

  Ours (ViT-S)  & Int8 
  & \textbf{49.5} & \textbf{54.9} & \textbf{52.2}
  & \textbf{27.4} & \textbf{34.3} & \textbf{30.9} \\

  \noalign{\vskip 2pt}
  \hline
  \end{tabular}
  }
\end{table}

%% file: table/crossdata_multimodal_accuracy.tex
\label{sub:cross_dataset_multimodal}

\begin{table}
  \centering
  \setlength\tabcolsep{6pt}
  \caption{Cross-dataset multimodal accuracy comparison between EdgeVL and our method.}
  \label{tab:multimodal_accuracy}
  \scalebox{0.95}{
  \begin{tabular}{l|c|cc}
  \hline
  \noalign{\vskip 2pt}
  Methods & Bits & NYUv2 (\%) $\uparrow$ & SUN-RGBD (\%) $\uparrow$ \\
  \noalign{\vskip 2pt}
  \hline
  \noalign{\vskip 2pt}

  EdgeVL (DAT-T)  \cite{cai2024self} & Int8 &  56.3 &  31.6 \\
  EdgeVL (Swin-T) \cite{cai2024self} & Int8 &  54.3 & 34.6\\
  EdgeVL (ViT-S)  \cite{cai2024self} & Int8 &  53.9 &  31.2\\

  \noalign{\vskip 2pt}
  \hline
  \noalign{\vskip 2pt}

  Ours (DAT-T)  & Int8 & \textbf{57.7} & \textbf{34.3} \\
  Ours (Swin-T) & Int8 & 52.5 & 30.0 \\
  Ours (ViT-S)  & Int8 & \textbf{56.6} & \textbf{35.0} \\

  \noalign{\vskip 2pt}
  \hline
  \end{tabular}
  }
\end{table}

%% file: table/loss_ablation.tex
\begin{table}[t]
\centering
\setlength{\tabcolsep}{6pt}
\renewcommand{\arraystretch}{1.15}
\small

\begin{minipage}[t]{0.57\linewidth}
\centering
\caption{Effect of adding loss components}
\label{tab:loss_ablation_style}
\begin{tabular}{l|cc}
\hline
Loss Terms & RGB & Non-RGB \\
\hline
$\mathcal{L}_{\text{rkd}}$ & 3.8 & 5.1 \\
$\mathcal{L}_{\text{rkd}} + \mathcal{L}_{\text{nce}}$ & 47.4 & 43.9 \\
$\mathcal{L}_{\text{rkd}} + \mathcal{L}_{\text{nce}} + \mathcal{L}_{\text{sup}}$ (full)
& \textbf{50.4} & \textbf{47.2} \\
\hline
\end{tabular}
\end{minipage}
\hfill
\begin{minipage}[t]{0.39\linewidth}
\centering
\caption{Robustness under lower bit-width.}
\label{tab:bit_robustness}
\begin{tabular}{c|cc}
\hline
Bits & RGB & Non-RGB \\
\hline
Int8 & 67.9 & \textbf{65.7} \\
Int6 & \textbf{68.6} & 65.3 \\
Int4 & 56.3 & 50.8 \\
\hline
\end{tabular}
\end{minipage}

\end{table}

%% file: table/ablation.tex
\begin{table}[t]
  \centering
  \setlength\tabcolsep{6pt}
  \caption{Effect of the proposed cross-modal adapter.}
  \label{tab:adapter_ablation}
  \scalebox{0.95}{
  \begin{tabular}{l|cc}
  \hline
  \noalign{\vskip 2pt}
  Methods & RGB (\%) $\uparrow$ & Non-RGB (\%) $\uparrow$ \\
  \noalign{\vskip 2pt}
  \hline
  \noalign{\vskip 2pt}

  Ours (DAT-T)              & \textbf{71.4} & \textbf{71.4} \\
  Ours w/o Adapter (DAT-T)  & 69.2 & 69.2 \\

  \noalign{\vskip 2pt}
  \hline
  \noalign{\vskip 2pt}

  Ours (Swin-T)             & \textbf{72.0} & \textbf{72.0} \\
  Ours w/o Adapter (Swin-T) & 70.7 & 70.7 \\

  \noalign{\vskip 2pt}
  \hline
  \noalign{\vskip 2pt}

  Ours (ViT-S)              & \textbf{70.8} & \textbf{70.8} \\
  Ours w/o Adapter (ViT-S)  & 69.5 & 69.5 \\

  \noalign{\vskip 2pt}
  \hline
  \end{tabular}
  }
\end{table}